\documentclass[letterpaper]{article} 
\usepackage{aaai25}  
\usepackage{times}  
\usepackage{helvet}  
\usepackage{courier}  
\usepackage[hyphens]{url}  
\usepackage{graphicx} 
\usepackage{natbib}  
\usepackage{caption} 
\usepackage{algorithm}
\usepackage{algorithmic}

\usepackage{newtxmath}  
\usepackage{amsmath} 
\usepackage{times}
\usepackage{latexsym}
\usepackage{booktabs} 
\usepackage{amsfonts} 

\usepackage{microtype}
\usepackage{multirow} 

\usepackage{caption}         

\usepackage{newfloat}
\usepackage{listings}
\DeclareCaptionStyle{ruled}{labelfont=normalfont,labelsep=colon,strut=off} 
\floatstyle{ruled}
\newfloat{listing}{tb}{lst}{}
\floatname{listing}{Listing}
\title{{DPCL-Diff: Temporal Knowledge Graph Reasoning Based on Graph Node Diffusion Model with Dual-Domain Periodic Contrastive Learning}}
\author{
    Yukun Cao, Lisheng Wang, Luobin Huang
}

\affiliations {
    Shanghai University of Electric Power, Shanghai, China\\ caoyukun@shiep.edu.cn, 2207992575@mail.shiep.edu.cn, 1877098981@mail.shiep.edu.cn
}

\usepackage{bibentry}

\begin{document}

\maketitle

\begin{abstract}
Temporal knowledge graph (TKG) reasoning that infers future missing facts is an essential and challenging task. Predicting future events typically relies on closely related historical facts, yielding more accurate results for repetitive or periodic events. However, for future events with sparse historical interactions, the effectiveness of this method, which focuses on leveraging high-frequency historical information, diminishes. Recently, the capabilities of diffusion models in image generation have opened new opportunities for TKG reasoning. Therefore, we propose a graph node diffusion model with dual-domain periodic contrastive learning (DPCL-Diff). Graph node diffusion model (GNDiff) introduces noise into sparsely related events to simulate new events, generating high-quality data that better conforms to the actual distribution. This generative mechanism significantly enhances the model's ability to reason about new events. Additionally, the dual-domain periodic contrastive learning (DPCL) maps periodic and non-periodic event entities to Poincaré and Euclidean spaces, leveraging their characteristics to distinguish similar periodic events effectively. Experimental results on four public datasets demonstrate that DPCL-Diff significantly outperforms state-of-the-art TKG models in event prediction, demonstrating our approach's effectiveness. This study also investigates the combined effectiveness of GNDiff and DPCL in TKG tasks.
\end{abstract}

%

\section{Introduction}

Knowledge graphs (KGs) store a large amount of information about the natural world and have shown great success in many downstream applications, such as natural language processing \cite{hu2022empowering, Wang-2020covid}, recommendation system \cite{xuan2023knowledge,yu2022graph}, and information retrieval \cite{zhou2022re}. Traditional KGs typically only contain static snapshots of facts, integrating facts (also known as events) in the form of static relational triples \((s, r, o)\), where \(s\) and \(o\) represent the subject and object entities, respectively, and \(r\) represents the relationship type. However, in the real world, knowledge evolves and constantly exhibits complex temporal dynamics \cite{wang2017knowledge,yin2016spatio}, which has inspired the construction and application of Temporal Knowledge Graph (TKG). TKG extends the previous static relational triples \((s, r, o)\) to quaternions \((s, r, o, t)\) with timestamps. Thus, the TKG consists of multiple snapshots where facts in the same snapshot appear simultaneously.

Existing studies have identified two main types of new events in TKG: periodic events that will recur \cite{liang2023learn,li2021temporal}, and events that have not occurred before but may occur in the future \cite{xu2023temporal}. For these new events, TKG reasoning offers fresh perspectives and insights for many downstream applications, such as event prediction \cite{zhou2023intensity}, political decision making \cite{zhang2022temporal}, dialogue generation \cite{hou2024selective}, and text generation \cite{li2022complex}. These applications have greatly stimulated intense interest in TKG. In this work, we focus on predicting new facts in future time, a task known as graph extrapolation \cite{chen2023temporal,li2021temporal}. Our goal is to predict the missing entities in the query \((s, p, ?, t)\) for future timestamps \(t\) that still need to be observed in the training set.

Most current work models the structural and temporal characteristics of TKG to capture the specific relationships and temporal dependencies between different events for future event prediction.Many studies \cite{liang2023learn,xu2023temporal,li2021temporal}can predict repetitive or periodic events by referring to the known historical events and distinguishing the different effects of periodic and non-periodic events on reasoning tasks through contrastive learning. However, in actual reasoning tasks, some periodic events share the same head entities and relations, differing only in their tail entities. This results in overly similar representations, making it difficult to distinguish them during reasoning. Additionally, In event-based TKG, new events that have never occurred account for about 40\% \cite{jager2018limits}. Due to the sparse traces of temporal interactions throughout the timeline of these new events, it is impossible to use reasoning methods based on high-frequency periodic events, resulting in poor reasoning performance for these types of events.

To address the above issue, we propose a diffusion model-based dual-doma in periodic contrast learning (DPCL-Diff) for temporal knowledge graph reasoning (TKG), using a Graph Node Diffusion model (GNDiff) to infer new events. GNDiff solves the sparse interaction traces for new events by adding noise to correlated sparse events of new entities on the timeline. This diffusion mechanism simulates the real-world occurrence of new events, generating high-quality data that closely align with the actual distribution through gradual correction during the denoising process. It also captures the event graph structure, enriching semantic representations and improving the model’s ability to reflect real-world dynamics. The diffusion model \cite{ho2020denoising,song2020denoising} is a state-of-the-art generative framework with proven strength in both image \cite{ramesh2022hierarchical,rombach2022high,saharia2022photorealistic} and text generation \cite{ou2024effective,gong2022diffuseq,strudel2022self}. This enhances new event inference and reasoning in TKGs. For tasks with large temporal interaction data, we propose a novel Dual-Domain Periodic Contrastive Learning (DPCL) method. Here, periodic and non-periodic event entities are mapped into Poincaré and Euclidean spaces for contrastive learning, respectively. The Poincaré space is used to better distinguish similar periodic entities, capturing entity relationships more effectively than Euclidean space, leading to more accurate link prediction (Balazevic, Allen, and Hospedales 2019; Zeb et al. 2024).

The main contributions of this thesis are summarized below:

\begin{itemize}
\item We propose a TKG model called DPCL-Diff. DPCL-Diff can reason not only about periodic events but also about new events through a diffusion generation mechanism.
\item GNDiff is introduced in the field of TKG. GNDiff can perform non-Markov decision chain-based graph node diffusion for new events, which generates a large amount of high-quality graph data related to new events.
\item We propose the DPCL method, where periodic and non-periodic event entities are mapped into Poincaré and Euclidean spaces, respectively, to more accurately capture the relationships between periodic event entities and distinguish similar ones.
\item We conducted experiments on four public datasets. The results show that DPCL-Diff outperforms the state-of-the-art TKG model in event prediction. Additionally, we purposefully explore the effectiveness of GNDiff and DPCL in the TKG task.
\end{itemize}

\begin{figure*}[t] 
    \centering
    \includegraphics[width=0.8\linewidth]{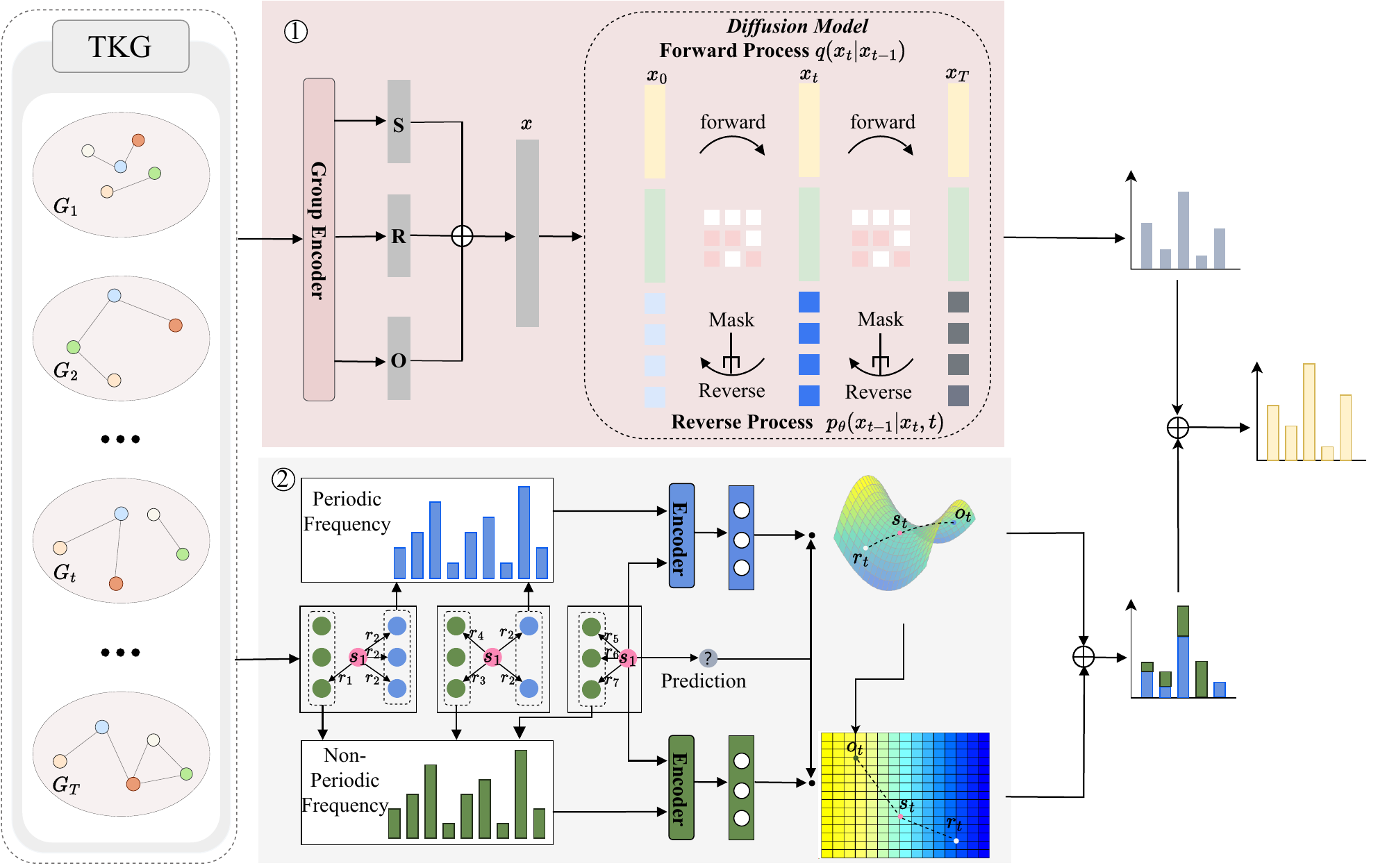}
    \caption{Overall architecture of DPCL-Diff. The first part is GNDiff, and the second part is DPCL.}
    \label{fig:DPCL-Diff}
\end{figure*}

\section{Related Work}

\subsection{Diffusion Model}

Diffusion models \cite{ho2020denoising} are latent variable models for continuous data, achieving state-of-the-art image and audio generation quality. For discrete data, prior work explored text diffusion in discrete spaces \cite{li2024few,lezama2022discrete,austin2021structured}. \citet{li2022diffusion} enables gradient-based control of text diffusion with continuous latent representations. \citet{ou2024effective} combines discrete diffusion and denoising to improve text generation quality and diversity in both unconditional and conditional tasks.

\subsection{Temporal Knowledge Graph Reasoning}

TKG reasoning involves interpolation and extrapolation \cite{chen2023temporal,li2021temporal}. Interpolation fills missing events within $t_{0}$ to $t_{n}$, while extrapolation predicts events beyond $t_{n}$. Interpolation models like TeAST \cite{li2023teast}, Re-Temp \cite{wang2023re}, and TempCaps \cite{fu2022tempcaps} address missing relationships but cannot predict future events. Extrapolation methods focus on future events: Know-Evolve \cite{trivedi2017know} captures nonlinear entity evolution but struggles with long-term dependencies, while CyGNet \cite{zhu2021learning} and CENET \cite{xu2023temporal} leverage repetitive patterns and contrastive learning. TARGAN \cite{xie2023targat} improves long-term modeling with attention mechanisms. Models like DynaQ \cite{liu2024reinforcement}, EvoGNN \cite{pareja2020evolvegcn}, and TempoR \cite{wu2020temp} enhance prediction but face challenges in capturing temporal evolution, semantic dependencies, and scalability.

\section{Method}

As shown in Fig.~\ref{fig:DPCL-Diff}, for new events, the DPCL-Diff model leverages GNDiff to inject noise into the sparse correlated events of the new event entity. This approach simulates the real-world mechanism of new events, generating numerous high-quality data samples. For other events, the DPCL method identifies highly correlated entities through contrastive learning and more accurately captures the relationships between periodic event entities using the properties of Poincaré space, thereby distinguishing similar periodic entities. In the following sections, we describe our proposed approach in detail.

\subsection{Preliminaries}
TKG \(G=\{G_0,G_1,...,G_t,...,G_T\}\) is a sequence of KGs, each \(G_{t}=\{\mathcal{E}, \mathcal{R}, \mathcal{T}\}\) contains facts that occur at timestamp \( t \). Let \(\mathcal{E}\), \(\mathcal{R}\) and \(\mathcal{T}\) denote a finite set of entities, relation types, and timestamps, respectively. Each fact is a quadruple \((s,r,o,t)\) , where \(s\in\mathcal{E}\) is the subject entity, \(o\in\mathcal{E}\) is the object entity, and \(r\in\mathcal{R}\) is the relation at timestamp \(t\) between \(s\) and \(o\) that occurs, the dimension is \(d\). \(\textbf{E}\in \mathbb{R}^{|\mathcal{E}|\times d}\) is the embedding of all entities, and its rows represent the embedding vectors of an entity, such as \(s\) and \(o\). Similarly, \(\textbf{R}\in \mathbb{R}^{|\mathcal{R}|\times d}\) is the embedding of all relation types. The TKG inference task aims to predict missing objects by answering a query \(q=(s,r,?,t)\) using a given history of TKG.

For a given query \(q=(s,r,?,t)\), we define the set of periodic events \(\mathcal{Q}_t^{s,r}\) as the set of events that have the same head entity relationship with the query event, and the corresponding set of periodic event entities as \(\mathcal{H}_t^{s,r}\) in the following equations:

\vspace{-10pt}
\begin{equation}
    \mathcal{Q}_t^{s,r}=\bigcup_{k<t}\{(s,r,o,k)\in\mathcal{G}_k\},
\end{equation}
\vspace{-10pt}

\vspace{-10pt}
\begin{equation}
    \mathcal{H}_t^{s,r}=\{o|(s,r,o,k)\in\mathcal{Q}_t^{s,r}\}.
\end{equation}
\vspace{-10pt}

Entities outside \(\mathcal{H}_t^{s,r}\) are non-periodic. If an event \((s,r,o,t)\) does not exist in \(\mathcal{Q}_t^{s,r}\), it is a new event that has never occurred and is handled by the GNDiff. For periodic event reasoning, DPCL is primarily used to predict target entities. The following sections detail these modules.

\subsection{Graph Node Diffusion Model}

In this section, we introduce GNDiff, demonstrating the effectiveness of diffusion models in TKGs. While diffusion models have been applied to discrete textual data, they are equally effective for discrete graph nodes in TKGs. Traditional methods, reliant on historical data, struggle with new events due to sparse data. In contrast, GNDiff uses a diffusion model to simulate event generation, producing high-quality samples for new event inference.

GNDiff includes a forward and a reverse diffusion process. For each quadruple \((s,r,o,t)\), the vector representations of \(s\), \(r\), and \(o\) are concatenated, forming a single vector as input to the graph node diffusion model. The sample data \(x_0 \sim q(x_0)\) is then obtained. Since TKGs are discrete, each element of \(x_t\) is a discrete random variable with \(K\) categories, where \(K = \left|\mathcal{V}\right|\) is the number of node types. The noise addition process is represented by a stack of heat vectors as shown in Equation 3:

\vspace{-10pt}
\begin{equation}
    q(x_t\mid x_{t-1})=\mathrm{Cat}(x_t;p=x_{t-1}Q_t),
\end{equation}
\vspace{-10pt}

where \(\mathrm{Cat}(\cdot)\) denotes that the noise addition process is based on the category distribution, and \({Q}_{t}\) is used to control the transition matrix of the node from one state to another:

\vspace{-10pt}
\begin{equation}
    [Q_t]_{i,j}=q(x_t=j\mid x_{t-1}=i).
\end{equation}
\vspace{-10pt}

The conversion relationship is shown in Equation 5:

\vspace{-10pt}
\begin{equation}
\begin{aligned}
&{q}(x_{t-1}\mid x_{t},x_{0})=\frac{q(x_{t}\mid x_{t-1},x_{0})q(x_{t-1}\mid x_{0})}{q(x_{t}\mid x_{0})}\\&=\mathrm{Cat}\Bigg(x_{t-1};p=\frac{x_{t}Q_{t}^{\top}\odot x_{0}\overline{Q}_{t-1}}{x_{0}\overline{Q}_{t}x_{t}^{\top}}\Bigg),
\end{aligned}
\end{equation}
\vspace{-10pt}

where \(\overline{Q}_{t}=Q_{1}Q_{2}\cdots Q_{t}\). Note that $\odot$ denotes elementwise multiplication, and division is performed rowwise. To incorporate the pre-trained language model into the denoising process, we introduce a specific "inactive" state as the absorbing state, allowing for the generation of samples during denoising. We use the mask mechanism to simulate the possible inactive state of the node. The transition matrix  adjustment process is shown in Equation 6:

\vspace{-10pt}
\begin{equation}
[Q_t]_{i,j}=
\begin{cases}
1 & \mathrm{if}\quad i=j=[\mathrm{M}], \\ 
\beta_t & \mathrm{if}\quad j=[\mathrm{M}],\ \mathrm{i}\neq[\mathrm{M}], \\
1-\beta_t & \mathrm{if}\quad i=j\neq[\mathrm{M}].
\end{cases}
\end{equation}
\vspace{-10pt}

Here, \([\mathrm{M}]\) denotes a node in a masked state (an inactive node), allowing nodes to transition naturally in and out of this state as the graph evolves, eventually converging to a smooth distribution \(q(x_T)\). This approach concentrates all probability mass on a sequence labelled with a mask. By using \(\overline{Q}_{t}=Q_{1}Q_{2}\cdots Q_{t}\), the transfer relation for multiple time steps can be derived. Samples are then generated through a backward diffusion process:

\vspace{-10pt}
\begin{equation}
p_\theta({x}_{0:T}) = p({x}_T) \prod\nolimits_{t=1}^T p_\theta({x}_{t-1} \mid {x}_t, t).
\end{equation}
\vspace{-10pt}

We optimize the inverse process by minimizing the variational lower bound to maximize the log-likelihood of the original data. This approach derives the optimization objective \(p_{\theta}({x}_{0:T})\) and controls the optimization through \(\mathcal{L}_{\mathrm{diff}}\):

\vspace{-10pt}
\begin{equation}
\begin{aligned}
&\mathcal{L}_{\mathrm{diff}} = \mathbb{E}_{q}[\mathrm{D}_{\mathrm{KL}}(q(x_{T} \mid x_{0}) \parallel p_{\theta}(x_{T}))] + \\
&\mathbb{E}_q\left[\sum\nolimits_{t=2}^T \mathrm{D}_{\mathrm{KL}}(q({x}_{t-1} \mid {x}_t, {x}_0) \parallel p_\theta({x}_{t-1} \mid {x}_t, t))\right] \\
&- \log p_{\theta}(x_{0} \mid x_{1}),
\end{aligned}
\end{equation}
\vspace{-10pt}

where \({E}_{q}\) denotes the expectation under a noisy data distribution. \(\mathrm{D}_\mathrm{KL}\) stands for Kullback-Leibler divergence.

To better control noise addition, GNDiff employs a noise schedule \cite{he2022diffusionbert} that measures noise at each step by masking and evenly distributing information. Tokens are assigned varying mask probabilities, ordered by informativeness into \(T\) stages. The forward process begins by masking the most informative tokens and ends with the least informative, simplifying the reverse process. Specifically, mask information is evenly distributed during the forward step, as shown in Equation 9:

\vspace{-10pt}
\begin{equation}
\sum\nolimits_{i=1}^n \overline{\alpha}_t^i H(x_0^i) = \left(1 - \frac{t}{T}\right) \sum\nolimits_{i=1}^n H(x_0^i),
\end{equation}
\vspace{-10pt}

where \(H\) represents entropy, quantifying the information content of a node. \(x_0^i\) denotes the \(i\)th token in the sequence and \(n\) is the sequence length. \(\overline{\alpha}_{t}^{i}=\prod_{j=1}^{t}(1-\beta_{j}^{i})\) is the probability that the node retains its original state at time step \(t\), i.e. \(x_t^i=x_0^i\). We expect that if \(H(x_{t}^{i})<H(x_{t}^{j})\), then  \(\bar{\alpha}_t^i>\bar{\alpha}_t^j\), so low-information tokens appear earlier in the reverse process than high-information ones. In practice, a token's entropy \(H(x)\) is computed as the negative logarithm of its frequency in the training corpus. Based on these properties, we construct \(\overline{\alpha}_{t}^{i}\) as follows:

\vspace{-10pt}
\begin{equation}
    \overline{\alpha}_{t}^{i}=1-\frac{t}{T}-G(t)\cdot\tilde{H}({x}_0^{i}),
\end{equation}
\vspace{-10pt}

\vspace{-10pt}
\begin{equation}
    G(t)=\mu\sin\frac{t\pi}T,
\end{equation}
\vspace{-10pt}

\vspace{-10pt}
\begin{equation}
    \tilde{H}({x}_0^i)=1-\frac{\sum_{j=1}^nH({x}_0^j)}{nH({x}_0^i)},
\end{equation}
\vspace{-10pt}

where \(G(t)\) controls the influence of information at time step \(t\), with its effect modulated by the hyperparameter \(\mu \).

\subsection{Dual-Domain Periodic Contrastive Learning}

Inspired by \citet{xu2023temporal}, we propose a dual-domain periodic contrastive learning method. Unlike the original approach, we use a dual-domain mapping strategy because we observed that specific queries involve multiple periodic events that are challenging to distinguish. To address this issue, we map entities from periodic and non-periodic events to Poincaré and Euclidean spaces, respectively, to identify highly correlated entities within each domain. By leveraging the spatial characteristics of Poincaré space, we more accurately capture the relationships between periodic entities and distinguish similar ones.

During data preprocessing, we analyze entity frequencies for a given query \(q=(s,r,?,t)\), as shown in Equation 13:

\vspace{-10pt}
\begin{equation}
\mathbf{Z}_t^{s,r}(o)=\lambda\cdot(\Phi_{\beta>0}-\Phi_{\beta=0}),
\end{equation}
\vspace{-10pt}

\vspace{-10pt}
\begin{equation}
\beta = \sum\nolimits_{(s,r,o_i,k)\in G_k}(o_i=o),
\end{equation}
\vspace{-10pt}

where the value of each entity frequency is constrained by the hyperparameter \(\lambda\). \(\Phi_{\beta}\) is an indicator function that returns 1 if \(\beta\) is true and 0 otherwise. \(\mathbf{Z}_t^{s,r}(o)>0\) indicates that quadruple \((s,r,o,t_k)\) is a periodic event bound to \(s\) and \(r\); otherwise, it is a non-periodic event. DPCL then learns dependencies in periodic and non-periodic events based on the input \(\mathbf{Z}_t^{s,r}(o)\). The distance between entities in Poincaré space is used as the final similarity score between an entity \(o_{i}\in \mathcal{E}\) and a query \(q\). The formula for calculating the Poincaré distance is provided as follows:

\vspace{-10pt}
\begin{equation}
    \mathbf{d}_{p}(s,o)=cosh^{-1}\Bigg(1+2\frac{\parallel s-o\parallel_2^2}{\left(1-\parallel s\parallel_2^2\right)\left(1-\parallel o\parallel_2^2\right)}\Bigg),
\end{equation}
\vspace{-10pt}

\vspace{-10pt}
\begin{equation}
    \mathbf{S}_{p}^{s,r}(o_{i})=tanh(\mathbf{W}_{p}{s}\oplus r+\mathbf{b}_{p})\mathbf{E}^{T}+\mathbf{Z}_{t}^{s,r}+\mathbf{d}_{p}(s,o_{i}),
\end{equation}
\vspace{-10pt}

where \(S_{p}^{s,r}\) denotes the periodic dependency score for different object entities, \({tanh}\) is the activation function, \(\oplus \) denotes the connection operator, and \(\mathbf{W}_{p}\in{R}^{d\times2d},\mathbf{b}_{p}\in{R}^{d\times2d}\) are the trainable parameters. We compute the initial similarity score between each entity \(o_{i}\in E\) and the query \(q\) using a linear layer with activation.

For non-periodic dependencies, we calculate the similarity scores \(d_{E}(s,o)=\parallel s- o\parallel\) using the following Euclidean distance formula to obtain the non-periodic dependency score \(S_{np}^{s,r}\):

\vspace{-10pt}
\begin{equation}
\mathbf{S}_{np}^{s,r}(o_{i})=tanh(\mathbf{W}_{np} {s} \oplus  r+\mathbf{b}_{np})\mathbf{E}^{T}-\mathbf{Z}_{t}^{s,r}+d_{{E}}(s,o_{i}).
\end{equation}
\vspace{-10pt}

The training objective of learning from periodic and non-periodic events is to minimize the loss \(\mathcal{L}_{\mathrm{ce}}\):

\vspace{-10pt}
\begin{equation}
\mathcal{L}_{\mathrm{ce}}=-\sum_{q\in I}\log\{\frac{exp(S_{p}^{s,r}(o_{i}))}{\sum\limits_{o_{j}\in\mathcal{E}}exp(S_{p}^{s,r}(o_{j}))}+\frac{exp(S_{np}^{s,r}(o_{i}))}{\sum\limits_{o_{j}\in\mathcal{E}}exp(S_{np}^{s,r}(o_{j}))}\}.
\end{equation}
\vspace{-10pt}

Here, \(o_{i}\) denotes the ground truth object entity for a given query \(q\). The goal of \(\mathcal{L}_{ce}\) is to distinguish the ground truth from other values by comparing each scalar in \(S_{p}^{s,r}\) and \(S_{np}^{s,r}\).

The model employs supervised contrastive learning to enhance its discriminative power by enforcing consistency within the same class and differentiation between different classes, controlled by the supervised contrastive loss \(\mathcal{L}_{sup}\):

\vspace{-10pt}
\begin{equation}
\mathcal{L}_{sup}=\sum_{q\in I}\frac{-1}{\mid P(i)\mid}\sum_{p\in P(i)}\log\frac{\exp(z_q\cdot z_p/\tau)}{\sum\exp(z_q\cdot z_a/\tau)},
\end{equation}
\vspace{-10pt}

where \(I\) is the set of positive samples, \(P(i)\) is the set of queries with the same label in the small batch, \(z_{q}\) denotes the embedding vector of query \(q\), \(z_{p}\) represents the embedding vector of queries sharing the same label as \(q\), \(z_{a}\) denotes the embedding vector of all non-\(q\) queries in the small batch.

\begin{table*}[t]
\centering
  \fontsize{9}{11}\selectfont
  \renewcommand{\arraystretch}{1} 
  \setlength{\tabcolsep}{3.5pt} 
  \begin{tabular}{c|cccc|cccc|cccc}
    \toprule
    \multirow{2}{*}{Method} & \multicolumn{4}{c|}{ICEWS18} & \multicolumn{4}{c|}{ICEWS14} & \multicolumn{4}{c}{WIKI} \\
    \cline{2-13}    
    & MRR & Hits@1 & Hits@3 & Hits@10 & MRR & Hits@1 & Hits@3 & Hits@10 & MRR & Hits@1 & Hits@3 & Hits@10 \\
    \midrule   
    ComplEx{(2016)} & 30.09 & 21.88 & 34.15 & 45.96 & 24.47 & 16.13 & 27.49 & 41.09 & 47.84 & 38.15 & 50.08 & 58.46 \\
    R-GCN{(2018)} & 23.19 & 16.36 & 25.34 & 36.48 & 26.31 & 18.23 & 30.43 & 45.34 & 37.57 & 28.15 & 39.66 & —— \\
    ConvE{(2018)} & 36.67 & 28.51 & 39.80 & 50.69 & 40.73 & 33.20 & 43.92 & 54.35 & 47.57 & 38.76 & 50.1 & 63.74 \\
    \hline
    RE-Net{(2019)} & 42.93 & 36.19 & 45.47 & 55.80 & 45.71 & 38.42 & 49.06 & 59.12  & 51.97 & 48.01 & 52.07 & 53.91 \\
    xERTE{(2020)} & 36.95 & 30.71 & 40.38 & 49.76 & 32.92 & 26.44 & 36.58 & 46.05 & 58.75 & 58.46 & 58.85 & 59.34 \\
    CyGNet{(2021)} & \underline{46.69} &  \textbf{\underline{40.58}} & \underline{49.82} & 57.14 & 48.63 & 41.77 & 52.50 & 60.29 & 45.5 & 50.48 & 50.79 & 68.95 \\
    EvoKG{(2022)} & 29.67 & 12.92 & 33.08 & \underline{58.32} & 18.30 & 6.30 & 19.43 & 39.37 & 50.66 & 12.21 & 63.84 & 64.73 \\
    RPC{(2023)} & 34.91 & 24.34 & 38.74 & 55.89 & 44.55 & 34.87 & 49.80 & 65.08 & —& —& —& — \\
    CENET{(2023)} & 43.72 & 37.91 & 45.65 & 54.79 & \underline{51.93} & \underline{48.24} & 52.81 & 59.33 & \underline{66.87} & \underline{66.77} & \underline{66.93} & 66.98 \\
    DHE-TKG{(2024)} & 29.23 & 19.15 & 33.31 & — & 40.02 & 30.13 & 44.99 & — & 51.20 & 57.47 & 69.25 & — \\
    RLGNet{(2024)} & 29.90 & 20.18 & 33.64 & 49.08 & 39.06 & 29.34 & 42.03 & 58.12 & 64.34 & 61.03 & 66.71 &  \textbf{\underline{69.51}}  \\
    HisRES{(2024)} & 37.69 & 26.46 & 42.75 & 59.70 & 50.48 & 39.57 & \underline{56.65} & \underline{71.09} & —& —& —& —\\
    \hline
    DPCL-Diff & \textbf{47.02} & 40.03 & \textbf{49.93} & \textbf{60.66} & \textbf{66.59} & \textbf{62.89} & \textbf{67.79} & \textbf{73.64} & \textbf{68.44} & \textbf{68.41} & \textbf{68.45} & {68.55}  \\
   \textbf{Improve(\%)}  & {7.55} {$\uparrow$} & {5.59} {$\uparrow$}  & {9.38}{$\uparrow$} & {10.71}{$\uparrow$} & {28.23} {$\uparrow$} & {30.37} {$\uparrow$} & {28.37} {$\uparrow$} & {24.12} {$\uparrow$} & {2.35} {$\uparrow$} & {2.46} {$\uparrow$}  & {2.26} {$\uparrow$}& {2.34} {$\uparrow$} \\
    \bottomrule
  \end{tabular}
  \caption{Experimental results of temporal link prediction on three datasets TKGs. The time-filtered MRR, H@1, H@3, and H@10 metrics are multiplied by 100. The best results are in bold, and the previous best results are underlined. \textbf{Improve} refers to the improvement of DPCL-Diff compared to our baseline CENET.}
\label{result1}
\end{table*}

\subsection{Training Details and Inference}


DPCL-Diff minimizes the loss function throughout model training:

\vspace{-10pt}
\begin{equation}
\mathcal{L}_{dpcl}=\alpha\cdot \mathcal{L}_{diff}+(1-\alpha)\cdot(\mathcal{L}_{ce}+\mathcal{L}_{sup}),
\end{equation}
\vspace{-10pt}

where \(\alpha \) is the equilibrium coefficient.


In GNDiff, the inference process closely mirrors the generation process, so candidate entity probability is computed via reverse diffusion.

\vspace{-10pt}
\begin{equation}
\mathbf{P}_{\text{diff}}(o_i \mid q) = \prod\nolimits_{t=1}^T p_\theta(x_{t-1} \mid x_t),
\end{equation}
\vspace{-10pt}

where \(o_{i}\) is the tail entity representation in the generated sample \(x_{t}\), and \(x_{t}\) represents the sample corresponding to the query \(q\).

For DPCL the probability distribution of entities is calculated based on the predicted missing objects:

\vspace{-10pt}
\begin{equation}
\mathbf{P}_{dpcl}(o_i\mid q)=\frac{\exp(s(o_i\mid q))}{\sum_{o_j\in\mathcal{E}}\exp(s(o_j\mid q))}.
\end{equation}
\vspace{-5pt}

The final probability \(\mathbf{P}(o_{i})\) is obtained by averaging the probabilities from the diffusion process and the DPCL prediction:

\vspace{-10pt}
\begin{equation}
\mathbf{P}(o_i\mid q)=\frac12\left(\mathbf{P}_{diff}(o_i\mid q)+\mathbf{P}_{dpcl}(o_i\mid q)\right).
\end{equation}
\vspace{-10pt}

DPCL-Diff will choose the object with the highest probability as the final prediction \(o_i\).

\section{Experiments}

\subsection{Experimental Setup}

\textbf{Datasets and Evaluation Metrics.} To evaluate the effectiveness of the proposed DPCL-Diff, we report the reasoning performance on four public datasets: ICEWS14 \cite{trivedi2017know}, ICEWS18 \cite{jin2019recurrent}, WIKI \cite{leblay2018deriving}, and YAGO \cite{mahdisoltani2013yago3}. We divided all datasets into three groups, roughly partitioning them into train (80\%), validation (10\%), and test (10\%) sets by timestamp. The partitioning of the ICEWS14 dataset differs slightly from the original dataset. Table \ref{dataset} shows more detailed information about the datasets. In our experiments, we chose the mean reversed rank (MRR) and the hit rate @{1,3,10} \cite{liang2023learn} as the metrics.

\begin{table}[t]
\fontsize{9}{11}\selectfont
\centering
\begin{tabular}{@{\hspace{2pt}}c@{\hspace{5pt}}c@{\hspace{5pt}}c@{\hspace{5pt}}c@{\hspace{5pt}}c@{\hspace{3pt}}c@{\hspace{3pt}}c@{\hspace{3pt}}}
\hline
Dataset & \#Ent & \#Re & \#Train & \#Valid & \#Test \\ \hline
ICEWS18  & 23,033        & 256    & 373,018   & 45,995      & 49,545 \\
ICEWS14     & 12,498 & 260 & 535,654 & 63,788 & 65,861 \\
WIKI     & 12,554 & 24  & 539,286 & 67,538 & 63,110 \\
YAGO      & 10,623 & 10  & 161,540 & 19,523 & 20,026 \\ \hline
\end{tabular}
\caption{Statistics of the datasets.}
\label{dataset}
\end{table}

\begin{table}[t]
\centering
  \fontsize{9}{11}\selectfont
  \renewcommand{\arraystretch}{1} 
  \setlength{\tabcolsep}{3.5pt} 
  \begin{tabular}{c|c c c c}
    \toprule
    \multirow{2}{*}{Method} & \multicolumn{4}{c}{YAGO} \\
    \cline{2-5}  
    & \centering MRR & Hits@1 & Hits@3 & Hits@10 \\
    \midrule   
    ComplEx{(2016)} & 61.29 & 54.88 & 62.28 & 69.42 \\
    R-GCN{(2018)} & 41.3 & 32.56 & 44.44 & —— \\
    ConvE{(2018)} & 62.32 & 56.19 & 63.97 & 70.44 \\
    \hline
    RE-Net{(2019)} & 65.16 & 63.29 & 65.63 & 68.08 \\
    xERTE{(2020)}  & 58.75 & 58.46 & 58.85 & 60.48 \\
    CyGNet{(2021)} & 63.47 & 64.26 & 65.71 & 68.95 \\
    EvoKG{(2022)}  & 55.11 & 54.37 & 81.38 & 92.73 \\
    CENET{(2023)} & 79.64 & \underline{78.65} & 79.91 & 81.42 \\
    DHE-TKG{(2024)} & 62.93 & 71.00 & 82.72 & — \\
    RLGNet{(2024)}  & \underline{80.17} & 76.52 & \underline{83.57} &  \textbf{\underline{84.96}} \\
    \hline
    DPCL-Diff   & \textbf{84.45} & \textbf{84.23} & \textbf{84.65} & 84.75 \\
   \textbf{Improve(\%)} & {6.04} {$\uparrow$}& {7.09} {$\uparrow$}  & {5.86} {$\uparrow$}& {4.09} {$\uparrow$}\\
    \bottomrule
  \end{tabular}
  \caption{Experimental results of temporal link prediction on YAGO.}
\label{result2}
\end{table}

\textbf{Baselines.} DPCL-Diff is compared with both static and temporal KG reasoning models. Static KG reasoning models, which do not consider temporal information, include ComplEx \cite{trouillon2016complex}, R-GCN \cite{schlichtkrull2018modeling} and ConvE \cite{dettmers2018convolutional}. In contrast, temporal KG reasoning models are designed to process temporal information and capture evolving patterns in TKG, such as RE-Net \cite{jin2019recurrent}, xERTE \cite{han2020targat}, CyGNet \cite{zhu2021learning}, EvoKG \cite{park2022evokg}, RPC \cite{liang2023learn}, CENET \cite{xu2023temporal}, DHE-TKG \cite{liu2024temporal}, RLGNet \cite{lv2024rlgnet}, and HisRES \cite{zhang2024historically}.

\begin{table*}[t]
  \fontsize{9}{11}\selectfont
  \renewcommand{\arraystretch}{1} 
  \setlength{\tabcolsep}{3.5pt} 
  \centering 
  \begin{tabular}{c|cccc|cccc}
    \toprule
    \multirow{2}{*}{Method} & \multicolumn{4}{c|}{ICEWS18} & \multicolumn{4}{c}{YAGO}\\ 
    \cline{2-9}  
    & MRR & Hits@1 & Hits@3 & Hits@10 & MRR & Hits@1 & Hits@3 & Hits@10  \\ 
    \midrule
    w/o- GNDif & 44.93 & 38.22 & 47.53 & 57.08 & 81.30 & 80.64 & 81.42 & 82.26  \\
    w/o- DPCL & 46.51 & 39.92 & 48.92 & 59.67 & 84.21 & 84.19 & 84.23 & 84.34 \\
    \midrule
    \multicolumn{1}{c|}{\textbf{DPCL-Diff}} & \textbf{47.02} & \textbf{40.03} & \textbf{49.93} & \textbf{60.66} & \textbf{84.45} & \textbf{84.23} & \textbf{84.65} & \textbf{84.75} \\
    \bottomrule
    \end{tabular}
    \caption{DPCL-Diff ablation studies on YAGO and ICEWS18 datasets.}
\label{ablation}
\end{table*}

\begin{table*}[t]
\centering
  \fontsize{9}{11}\selectfont
  \renewcommand{\arraystretch}{1} 
  \setlength{\tabcolsep}{3.5pt} 
  \begin{tabular}{c|cccc|cccc|cccc}
    \toprule
  \multirow{2}{*}{Method} & \multicolumn{4}{c|}{ICEWS14(new events)} & \multicolumn{4}{c|}{ICEWS18(new events)} & \multicolumn{4}{c}{WIKI(new events)} \\ 
    \cline{2-13}    
    & MRR & Hits@1 & Hits@3 & Hits@10 & MRR & Hits@1 & Hits@3 & Hits@10 & MRR & Hits@1 & Hits@3 & Hits@10 \\
    \midrule   
    GNDiff &47.36 & 41.56 & 49.04 & 58.72 & 37.46 & 31.64 & 39.71 & 48.42  & 62.51 & 61.97 & 62.77 & 63.37 \\
    DPCL & 42.37 & 36.41 & 44.78 & 53.03 & 33.93 & 29.47 & 35.21 & 42.18 & 57.16 & 56.81 & 57.28 & 57.58 \\
    \bottomrule
  \end{tabular}
\caption{Performance of Graph Diffusion Models on New Events in ICEWS14 and ICEWS18 and WIKI.}
\label{ICEWS14}
\end{table*}

\begin{table}[]
\setlength{\tabcolsep}{3.5pt} 
\fontsize{9}{11}\selectfont
\centering
\begin{tabular}{c|cccc}
\toprule
\multirow{2}{*}{Method} & \multicolumn{4}{c}{YAGO(new events)} \\ 
\cline{2-5}  
& MRR & Hits@1 & Hits@3 & Hits@10 \\ 
\midrule
GNDiff & 79.49 & 79.44 & 79.48 & 79.54 \\
DPCL & 71.88 & 71.78 & 71.89 & 71.98 \\
\bottomrule
\end{tabular}
\caption{Performance of Graph Diffusion Models on New Events in YAGO.}
\label{YAGO}
\end{table}

\begin{table}[h]
\fontsize{9}{11}\selectfont
\centering
\begin{tabular}
{@{\hspace{2pt}}c@{\hspace{5pt}}c@{\hspace{5pt}}c@{\hspace{3pt}}c@{\hspace{3pt}}c@{\hspace{3pt}}}
\hline
New Events Dataset & \#Train &\#Valid & \#Test \\ \hline
ICEWS18  & 221,000 & 31,407 & 33,211 \\
ICEWS14  & 275,762 & 35,744 & 34,641 \\
WIKI  & 113,629 & 30,317 & 28,811 \\
YAGO   & 44,949  & 10,704 & 10,985 \\ \hline
\end{tabular}
\caption{Statistics of New Events Datasets.}
\label{Dataset1}
\end{table}

\textbf{Implementation Details.} 
For all datasets, DPCL uses an embedding dimension of 200, while GNDiff uses 128. We set the batch size to 64, the learning rate to 0.001, and used the Adam optimizer. The training calendar element limit for \( L \) is 30, and the comparison learning calendar element limit for the second stage is 20. The value of the hyperparameter \( \alpha \) is set to 0.2, and \( \lambda \) is set to 2. Additionally, \( \tau \) is the temperature parameter set to 0.1 in the supervised contrastive loss \( \mathcal{L}_{sup} \). For baseline settings, we use their recommended configuration. All experiments were implemented on servers with RTX 3090 GPUs equipped with Intel Xeon Gold 6330 CPUs.

\subsection{Main Results}

We compared the experimental results in Table \ref{result1} and Table \ref{result2}, presenting the MRR and Hits@1/3/10 metrics of DPCL-Diff across four datasets. DPCL-Diff consistently outperforms 12 baseline models, with a notable \textbf{30.37\%(Hits@1)} improvement over CENET on ICEWS14. This substantial gain is not solely attributable to the effective data generation of GNDiff for handling unseen events but also to its ability to generalize on periodic events. ICEWS14 high proportion of new events and large data volume amplify GNDiff impact, while the pre-trained language model (PLM) further enhances contextual representation. Additionally, DPCL-Diff's ability to model structural and semantic relationships across events improves performance on both new and periodic events, with the sensitivity of the Hits@1 metric to ranking adjustments further emphasizing the improvement.
 
On the YAGO and WIKI datasets, DPCL-Diff also achieves strong overall performance, with significantly higher Hits@1 compared to other models, even though Hits@10 on YAGO is slightly lower than RLGNet. The model’s ability to capture relationships between periodic events enhances prediction accuracy despite the lower proportion of new events and a higher proportion of periodic events, which limit GNDiff’s direct impact. Furthermore, the experiments were conducted with a relatively small batch size of 64, potentially constraining the model’s performance. Increasing the batch size to 512 or 1024 could further enhance its capacity to capture contextual and relational information, especially in datasets with more complex event patterns.

\subsection{Ablation Study}
To investigate the effectiveness of the modules in DPCL-Diff, we verified the effectiveness of GNDiff and DPCL through ablation experiments. The ablation results are shown in Table \ref{ablation}. Here, "w/o-GN Diff" denotes the removal of the graph node diffusion model, and "w/o-DPCL" denotes the removal of the Poincaré mapping, which maps both positive and negative samples into the Euclidean space for contrast learning. These results illustrate the effectiveness of our proposed GNDiff and DPCL. The performance degradation is particularly noticeable after the removal of GNDiff. This demonstrates that generating a large number of high-quality data related to the predicted event through GNDiff significantly affects event prediction. DPCL leverages the properties of Poincaré space to better capture the relationships between periodic entities and distinguish similar periodic events, thereby enhancing inference accuracy. 

\subsection{Effectiveness Analysis of Graph Diffusion in New Events}

To evaluate the effectiveness of graph diffusion models on new events, we curated datasets from four sources by selecting the first occurrences of distinct events, thereby focusing exclusively on new data. We extracted datasets based on the first appearance of all events that occurred, ensuring that only new and unique occurrences were included, while subsequent appearances of the same events were excluded. Detailed dataset statistics are provided in Table \ref{Dataset1}, which supports our methodological approach. The performance comparisons, detailed in Table \ref{ICEWS14} and Table \ref{YAGO}, show that the GNDiff model consistently outperforms the DPCL model across all metrics and datasets. 
This robust performance highlights the model's capability to adeptly handle dynamic, unseen data, proving GNDiff's exceptional utility in environments that require the continuous integration and processing of new information. 

\begin{table}[t]
  \fontsize{9}{11}\selectfont
  \renewcommand{\arraystretch}{1} 
  \setlength{\tabcolsep}{3.5pt} 
  \centering
  \begin{tabular}{c|cccc}
    \toprule
    \multirow{2}{*}{Method} & \multicolumn{4}{c}{YAGO} \\ 
    \cline{2-5}  
    & MRR & Hits@1 & Hits@5 & Hits@10 \\ 
    \midrule
    Euc/Hyp & 81.20 & 80.60 & 81.26 & 82.08 \\
    Hyp/Hyp & 81.23 & 80.62 & 81.30 & 82.14 \\
    Euc/Euc & 79.64 & 78.65 & 79.91 & 81.42 \\
    \midrule
    \multicolumn{1}{c|}{\textbf{Hyp/Euc}} & \textbf{81.30} & \textbf{80.64} & \textbf{81.42} & \textbf{82.26} \\
    \bottomrule
  \end{tabular}
    \caption{Experimental results of DPCL under different spatial mapping strategies}
\label{strategy}
\end{table}

\subsection{Dual Domain Mapping Strategy Analysis}

In this study, we examined the effects of different spatial mapping strategies on DPCL's learning performance. Table \ref{strategy} presents the results on the YAGO dataset without GNDiff. Euc/Hyp denotes that periodic and non-periodic event entities are mapped to Euclidean and Poincaré spaces, respectively, for comparative learning, Hyp/ Euc denotes that the two types of entities are mapped to Poincaré and Euclidean space, Hyp/Hyp denotes that both types of entities are mapped to Poincaré space. Euc / Euc denotes that the two types of entities are mapped to Euclidean space, respectively. The experimental results show that the Hyp/Euc method is the most effective, while the Euc/Euc method is the least effective. This suggests that mapping periodic and non-periodic event entities to separate spaces can improve the modelling of comparative learning. Specifically, compared to Euclidean space, the geometric properties of the Poincaré space allow for a more accurate capture of the relationships between periodic entities and distinguish similar ones, leading to better link prediction results.

\subsection{Dual Domain Embedded Visual Analytics}

\begin{figure}[tb]	
	\begin{minipage}{0.49\linewidth}
		\vspace{1pt}

		\centerline{\includegraphics[width=\textwidth]{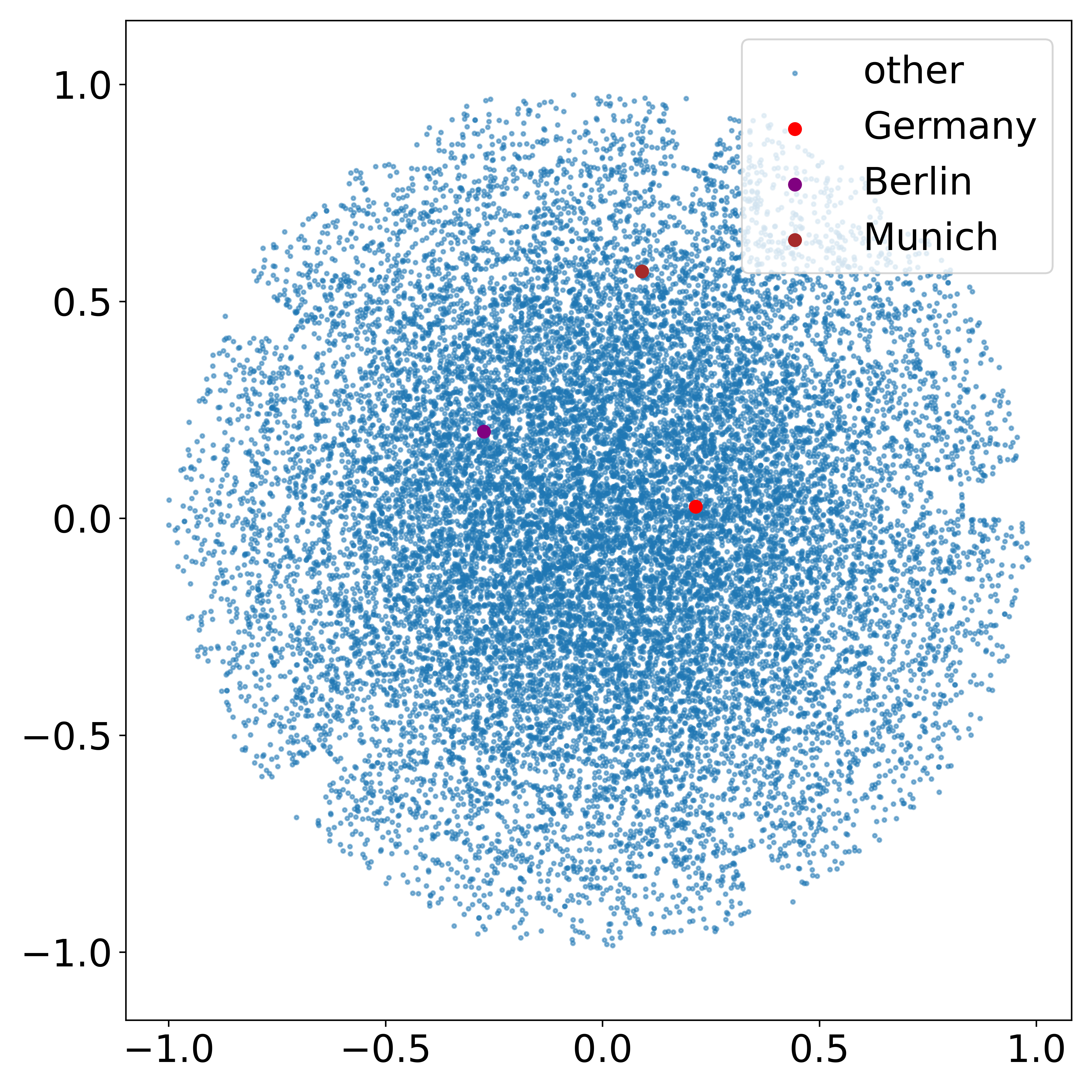}}    
	\end{minipage}
	\begin{minipage}{0.49\linewidth}
		\vspace{1pt}
		\centerline{\includegraphics[width=\textwidth]{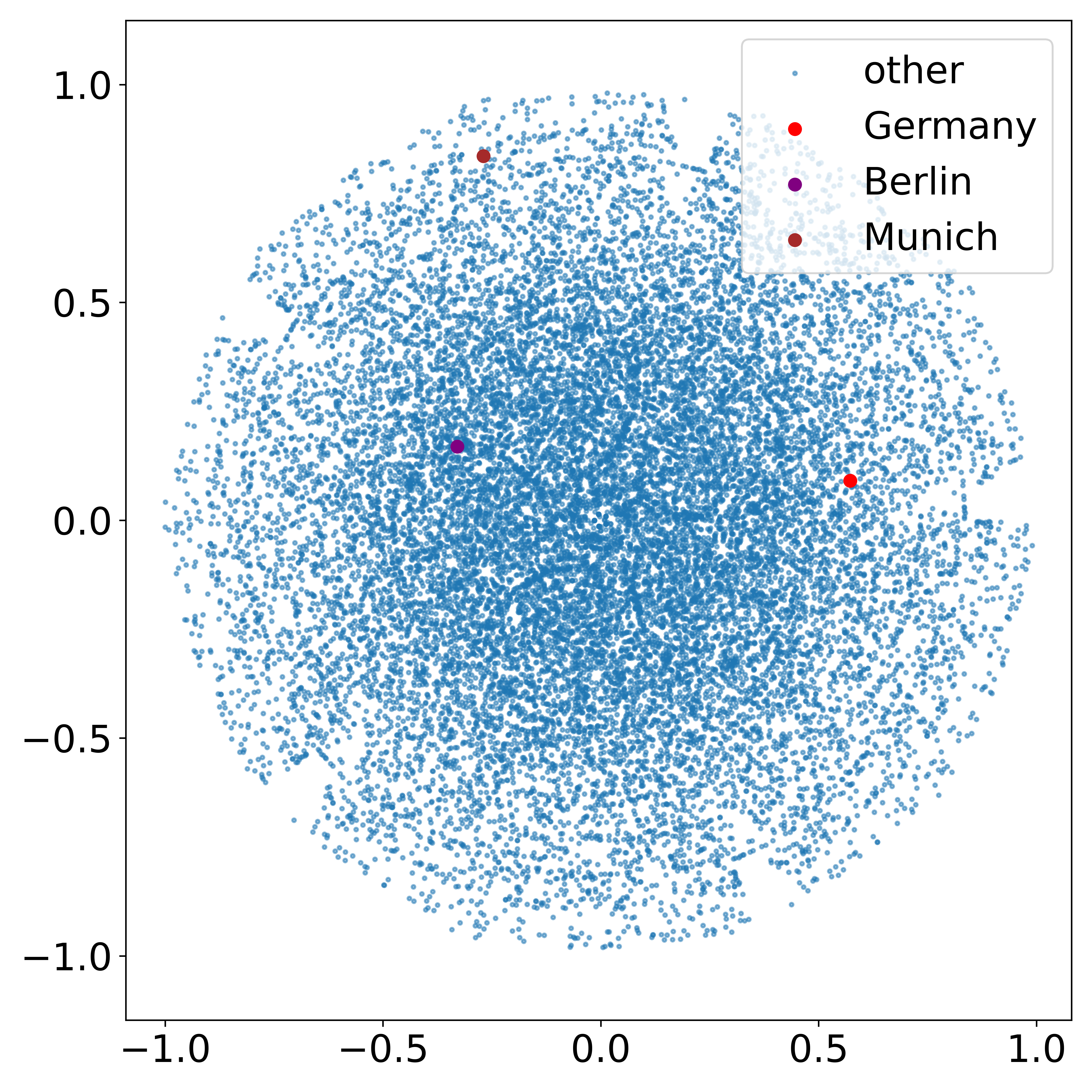}}
	\end{minipage}
 
	\caption{Illustration of 2D entity embedding learned by Euc (left) and Hyp (right) on YAGO.}
	\label{embedded}
\end{figure}

In this section, we visualize entity pairs—(Germany, Berlin) and (Germany, Munich)—with the same relation, embedded in both Euclidean and Poincaré spaces, as shown in Fig.~\ref{embedded}. In Euclidean space, the relationships between Germany and the two cities are not well differentiated. In contrast, Poincaré space provides a clearer distinction, better reflecting the relationships between the entities. By leveraging the geometric properties of Poincaré space, we improve the model's ability to capture entity relationships, better distinguish similar entities, thereby enhancing prediction accuracy.

\subsection{Graph Node Diffusion Generation Process}

To address the issue of insufficient interaction traces for new events on the timeline, GNDiff generates data samples by simulating new events using a diffusion mechanism. Table \ref{node} illustrates this process for GNDiff. Since graph node data differs from text data, we first obtain node embedding vectors and then apply the diffusion mechanism. Each node is transformed into a 128-dimensional vector, but for simplicity, we use a 4-dimensional vector to demonstrate the process.

The process begins with a fully masked state and progressively reveals information about entities and relationships. Initially (t = 0), all information is masked. For instance, even though entity 1 (United States) and the relation (sanctioned) are known, they are masked, requiring the model to predict entity 2. By step three (t = 3), the model predicts entity 2 based on frequency information, generating a vector [0.23, 0.37, 0.52, ?]. In step four (t = 4), the model adjusts this vector to [0.18, 0.29, 0.71, ?]. Finally, the model generates the complete vector [0.18, 0.29, 0.71, 0.86], mapping it to the entity "Russia," thus completing the generation process. This iterative adjustment enhances the quality of samples, improving the model's accuracy and robustness in predicting new events.

\subsection{Hyper-parameter Analysis}

\begin{figure}[tb]
	
	\begin{minipage}{0.48\linewidth}
		\vspace{3pt}
        
		\centerline{\includegraphics[width=\textwidth]{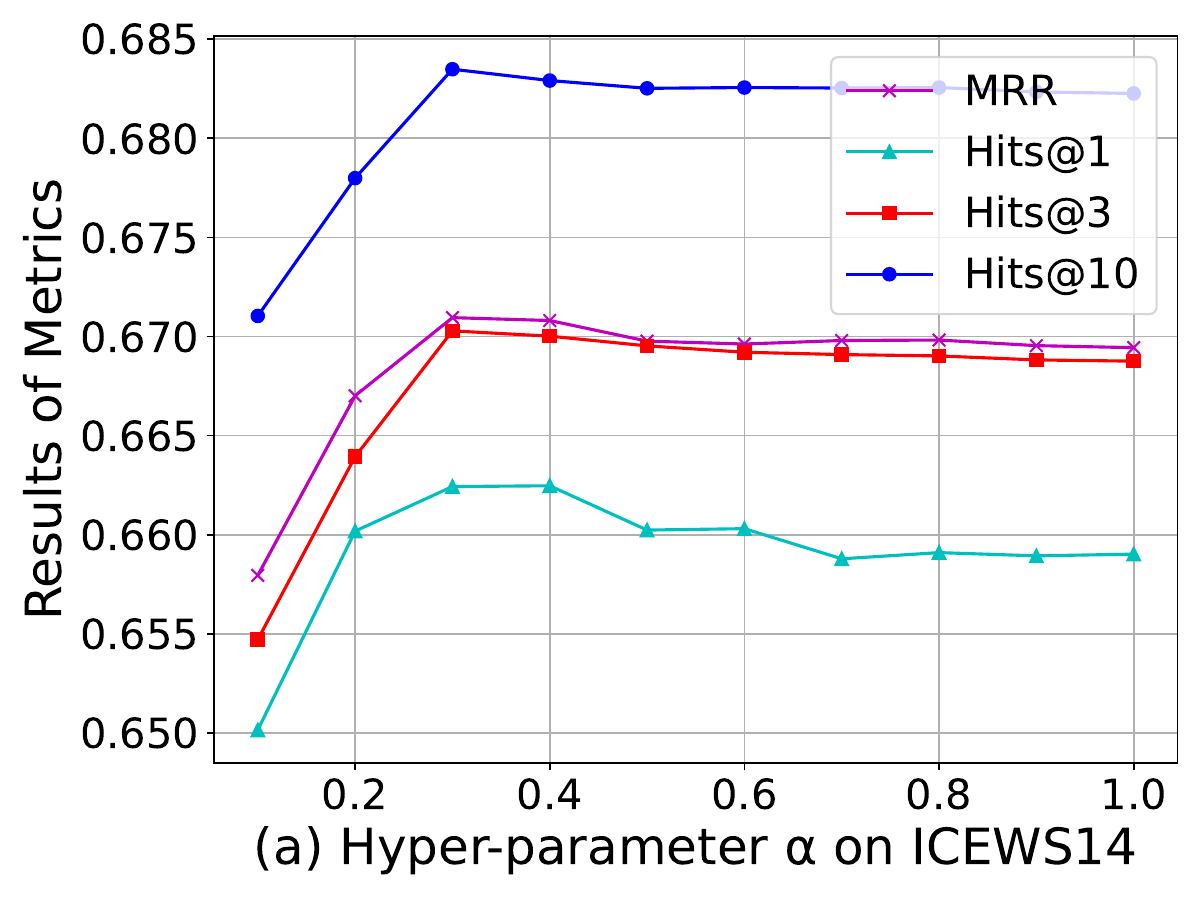}}
        
	\end{minipage}
	\begin{minipage}{0.48\linewidth}
		\vspace{3pt}
		\centerline{\includegraphics[width=\textwidth]{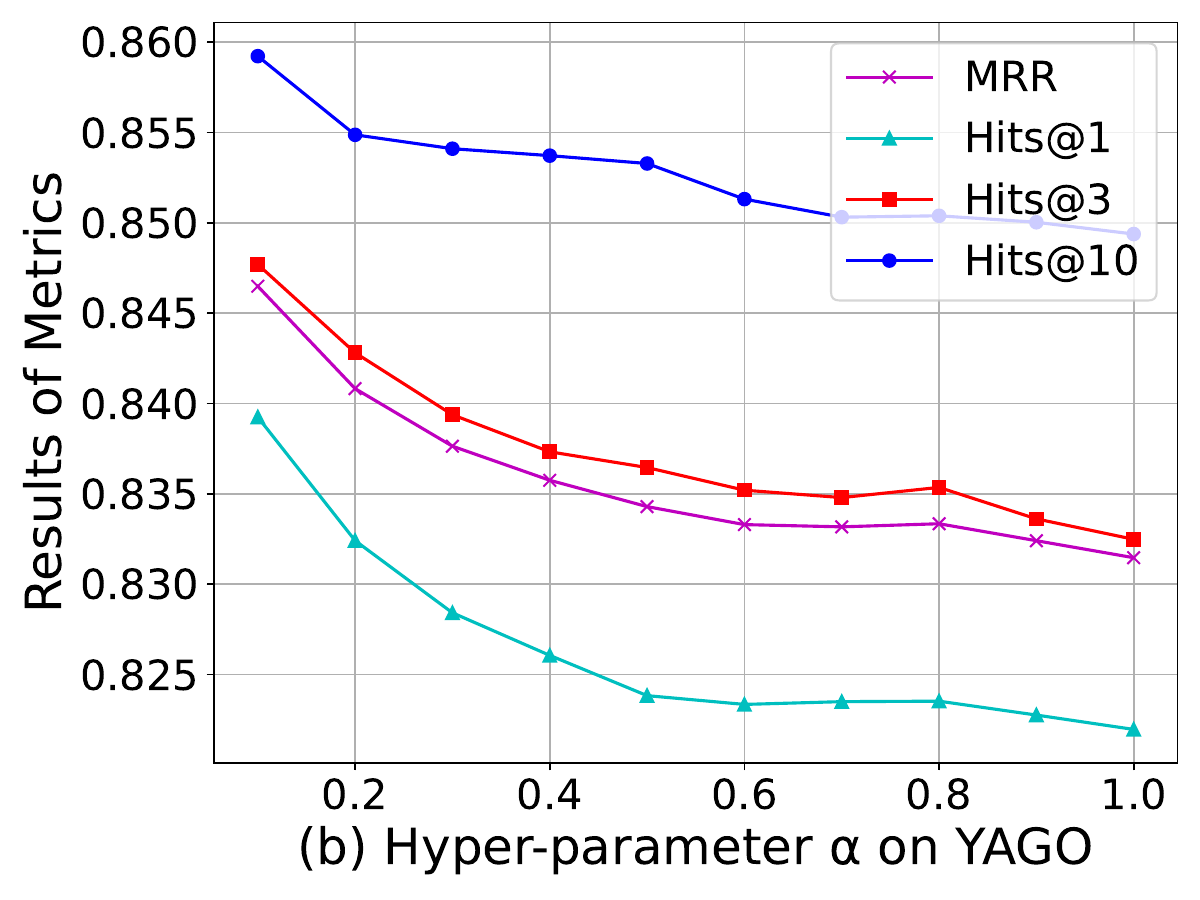}}
        
	\end{minipage}

	\begin{minipage}{0.48\linewidth}
		\vspace{3pt}
        
		\centerline{\includegraphics[width=\textwidth]{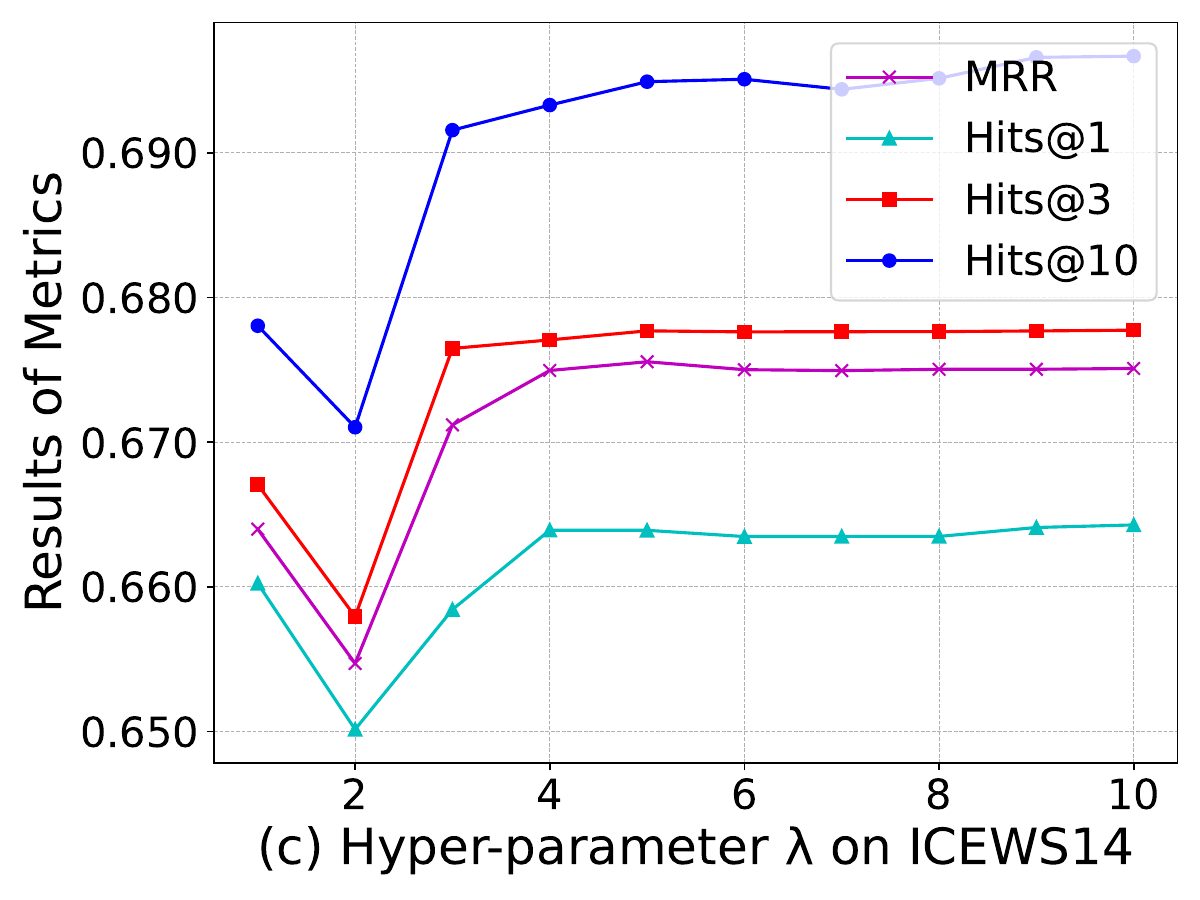}}
        
	\end{minipage}
        \begin{minipage}{0.48\linewidth}
		\vspace{3pt}
		\centerline{\includegraphics[width=\textwidth]{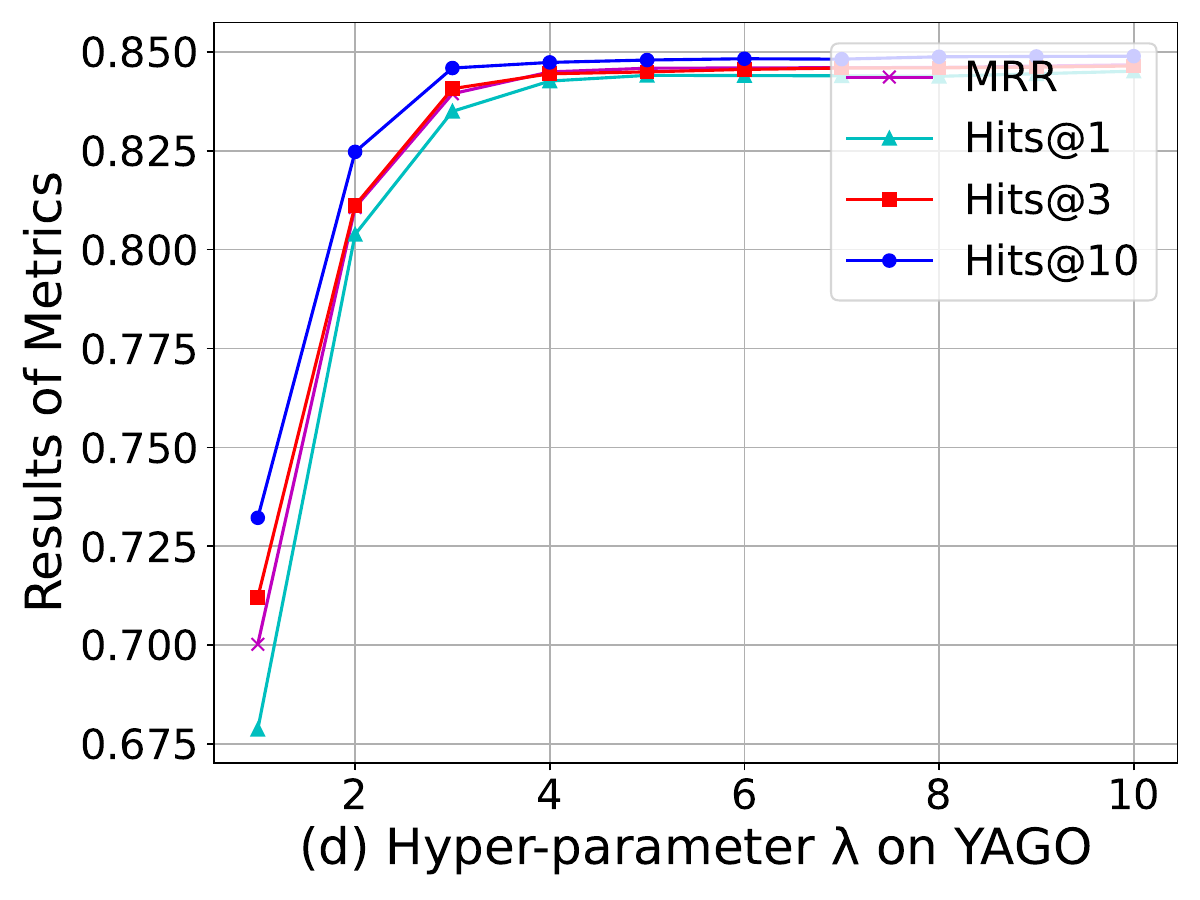}}
        
	\end{minipage}

        \caption{Results of hyper-parameters \(\alpha \) and $\lambda$ of DPCL-Diff on ICEWS14 and YAGO.}
	\label{Hyper-parameter}
\end{figure}

DPCL-Diff involves two key hyperparameters, $\alpha$ and $\lambda$, whose impact on performance is analyzed for the ICEWS14 and YAGO datasets in Fig.~\ref{Hyper-parameter}. $\alpha$ balances $L_{\text{diff}}$ and $L_{\text{dpcl}}$. On ICEWS14, $\alpha=0.3$ enhances temporal interaction modeling with minimal sensitivity to changes. For YAGO, a lower $\alpha$ is more effective, highlighting the importance of dual-domain contrastive learning for periodic events. For $\lambda$, after fixing $\alpha$, $\lambda=4$ optimizes performance on ICEWS14 by balancing temporal variations and preventing overfitting. On YAGO, progressively increasing $\lambda$ improves performance, handling the dataset’s complex structure. These results demonstrate DPCL-Diff's robustness and ability to maintain high performance with stable hyperparameter settings, effectively balancing diffusion and contrastive learning across datasets.

\begin{table}[tb]
  \small
  \renewcommand{\arraystretch}{1.5} 
  \setlength{\tabcolsep}{3pt} 
  \centering
  \begin{tabular}{l l l l}
  \hline
    \toprule
    \textbf{step} & \textbf{Entity 1 } & \textbf{Relation} & \textbf{Entity 2 } \\ 
    \midrule
    t = 0 & [MASK] & [MASK] & [MASK] \\
    t = 1 & United States & [MASK] & [MASK] \\
    t = 2 & United States & sanctioned & [MASK] \\
    t = 3 & United States & sanctioned & [0.23, 0.37, 0.52, ?] \\
    t = 4 & United States & sanctioned & [0.18, 0.29, 0.71, ?] \\
    t = 4 & United States & sanctioned & [0.18, 0.29, 0.71, 0.86] \\
    t = 5 & United States & sanctioned & Russia \\
    \bottomrule
  \end{tabular}
      \caption{ Example of graphical node diffusion model generation process}
\label{node}
\end{table}

\section{Conclusion}

In this paper, we propose a temporal knowledge graph inference method based on a graph node diffusion model with dual-domain periodic contrastive learning. This method aims to alleviate the prediction difficulties caused by the limited interactions of new events on the timeline. We enhance new event reasoning by generating graph data through graph node diffusion, where noise is introduced and adjusted to simulate real-world event mechanisms, producing samples that better reflect actual distributions. Additionally, the performance of the temporal knowledge reasoning task is enhanced by mapping event entities to Poincaré space and Euclidean space through dual-domain periodic contrastive learning, effectively distinguishing between periodic and non-periodic events. Extensive experiments demonstrate the reasoning advantages and effectiveness of DPCL-Diff on public temporal knowledge graph datasets.

\section{Acknowledgments}

This work was supported by the National Natural Science Foundation of China (No. 62402307, No. U23B2021).

\newpage
\bibliography{aaai25}

\begin{thebibliography}{49}
\providecommand{\natexlab}[1]{#1}

\bibitem[{Austin et~al.(2021)Austin, Johnson, Ho, Tarlow, and Van Den~Berg}]{austin2021structured}
Austin, J.; Johnson, D.~D.; Ho, J.; Tarlow, D.; and Van Den~Berg, R. 2021.
\newblock Structured denoising diffusion models in discrete state-spaces.
\newblock \emph{Advances in Neural Information Processing Systems}, 34: 17981--17993.

\bibitem[{Chen et~al.(2023)Chen, Xu, Su, Huang, and Dou}]{chen2023temporal}
Chen, Z.; Xu, C.; Su, F.; Huang, Z.; and Dou, Y. 2023.
\newblock Temporal Extrapolation and Knowledge Transfer for Lifelong Temporal Knowledge Graph Reasoning.
\newblock In \emph{Findings of the Association for Computational Linguistics: EMNLP 2023}, 6736--6746.

\bibitem[{Dettmers et~al.(2018)Dettmers, Minervini, Stenetorp, and Riedel}]{dettmers2018convolutional}
Dettmers, T.; Minervini, P.; Stenetorp, P.; and Riedel, S. 2018.
\newblock Convolutional 2d knowledge graph embeddings.
\newblock In \emph{Proceedings of the AAAI conference on artificial intelligence}, volume~32.

\bibitem[{Fu et~al.(2022)Fu, Meng, Han, Ding, Ma, Schubert, Tresp, and Wattenhofer}]{fu2022tempcaps}
Fu, G.; Meng, Z.; Han, Z.; Ding, Z.; Ma, Y.; Schubert, M.; Tresp, V.; and Wattenhofer, R. 2022.
\newblock TempCaps: a capsule network-based embedding model for temporal knowledge graph completion.
\newblock In \emph{Proceedings of the Sixth Workshop on Structured Prediction for NLP}, 22--31. Association for Computational Linguistics.

\bibitem[{Gong et~al.(2022)Gong, Li, Feng, Wu, and Kong}]{gong2022diffuseq}
Gong, S.; Li, M.; Feng, J.; Wu, Z.; and Kong, L. 2022.
\newblock Diffuseq: Sequence to sequence text generation with diffusion models.
\newblock \emph{arXiv preprint arXiv:2210.08933}.

\bibitem[{He et~al.(2022)He, Sun, Wang, Huang, and Qiu}]{he2022diffusionbert}
He, Z.; Sun, T.; Wang, K.; Huang, X.; and Qiu, X. 2022.
\newblock Diffusionbert: Improving generative masked language models with diffusion models.
\newblock \emph{arXiv preprint arXiv:2211.15029}.

\bibitem[{Ho, Jain, and Abbeel(2020)}]{ho2020denoising}
Ho, J.; Jain, A.; and Abbeel, P. 2020.
\newblock Denoising diffusion probabilistic models.
\newblock \emph{Advances in neural information processing systems}, 33: 6840--6851.

\bibitem[{Hou et~al.(2024)Hou, Jin, Li, Bai, Guo, and Cheng}]{hou2024selective}
Hou, Z.; Jin, X.; Li, Z.; Bai, L.; Guo, J.; and Cheng, X. 2024.
\newblock Selective Temporal Knowledge Graph Reasoning.
\newblock \emph{arXiv preprint arXiv:2404.01695}.

\bibitem[{Hu et~al.(2022)Hu, Xu, Yu, Wang, Yang, Zhu, Chang, and Sun}]{hu2022empowering}
Hu, Z.; Xu, Y.; Yu, W.; Wang, S.; Yang, Z.; Zhu, C.; Chang, K.-W.; and Sun, Y. 2022.
\newblock Empowering language models with knowledge graph reasoning for question answering.
\newblock \emph{arXiv preprint arXiv:2211.08380}.

\bibitem[{J{\"a}ger(2018)}]{jager2018limits}
J{\"a}ger, K. 2018.
\newblock The limits of studying networks via event data: Evidence from the ICEWS dataset.
\newblock \emph{Journal of Global Security Studies}, 3(4): 498--511.

\bibitem[{Jin et~al.(2019)Jin, Qu, Jin, and Ren}]{jin2019recurrent}
Jin, W.; Qu, M.; Jin, X.; and Ren, X. 2019.
\newblock Recurrent event network: Autoregressive structure inference over temporal knowledge graphs.
\newblock \emph{arXiv preprint arXiv:1904.05530}.

\bibitem[{Leblay and Chekol(2018)}]{leblay2018deriving}
Leblay, J.; and Chekol, M.~W. 2018.
\newblock Deriving validity time in knowledge graph.
\newblock In \emph{Companion proceedings of the the web conference 2018}, 1771--1776.

\bibitem[{Lezama et~al.(2022)Lezama, Salimans, Jiang, Chang, Ho, and Essa}]{lezama2022discrete}
Lezama, J.; Salimans, T.; Jiang, L.; Chang, H.; Ho, J.; and Essa, I. 2022.
\newblock Discrete predictor-corrector diffusion models for image synthesis.
\newblock In \emph{The Eleventh International Conference on Learning Representations}.

\bibitem[{Li et~al.(2024)Li, Gao, Zhu, Yin, Cao, Jiang, and Xu}]{li2024few}
Li, B.; Gao, Z.; Zhu, Y.; Yin, K.; Cao, H.; Jiang, D.; and Xu, L. 2024.
\newblock Few-shot Temporal Pruning Accelerates Diffusion Models for Text Generation.
\newblock In \emph{Proceedings of the 2024 Joint International Conference on Computational Linguistics, Language Resources and Evaluation (LREC-COLING 2024)}, 7259--7269.

\bibitem[{Li, Su, and Gao(2023)}]{li2023teast}
Li, J.; Su, X.; and Gao, G. 2023.
\newblock Teast: Temporal knowledge graph embedding via archimedean spiral timeline.
\newblock In \emph{Proceedings of the 61st Annual Meeting of the Association for Computational Linguistics (Volume 1: Long Papers)}, 15460--15474.

\bibitem[{Li et~al.(2022{\natexlab{a}})Li, Thickstun, Gulrajani, Liang, and Hashimoto}]{li2022diffusion}
Li, X.; Thickstun, J.; Gulrajani, I.; Liang, P.~S.; and Hashimoto, T.~B. 2022{\natexlab{a}}.
\newblock Diffusion-lm improves controllable text generation.
\newblock \emph{Advances in Neural Information Processing Systems}, 35: 4328--4343.

\bibitem[{Li et~al.(2022{\natexlab{b}})Li, Guan, Jin, Peng, Lyu, Zhu, Bai, Li, Guo, and Cheng}]{li2022complex}
Li, Z.; Guan, S.; Jin, X.; Peng, W.; Lyu, Y.; Zhu, Y.; Bai, L.; Li, W.; Guo, J.; and Cheng, X. 2022{\natexlab{b}}.
\newblock Complex evolutional pattern learning for temporal knowledge graph reasoning.
\newblock \emph{arXiv preprint arXiv:2203.07782}.

\bibitem[{Li et~al.(2021)Li, Jin, Li, Guan, Guo, Shen, Wang, and Cheng}]{li2021temporal}
Li, Z.; Jin, X.; Li, W.; Guan, S.; Guo, J.; Shen, H.; Wang, Y.; and Cheng, X. 2021.
\newblock Temporal knowledge graph reasoning based on evolutional representation learning.
\newblock In \emph{Proceedings of the 44th international ACM SIGIR conference on research and development in information retrieval}, 408--417.

\bibitem[{Liang et~al.(2023)Liang, Meng, Liu, Liu, Tu, Wang, Zhou, and Liu}]{liang2023learn}
Liang, K.; Meng, L.; Liu, M.; Liu, Y.; Tu, W.; Wang, S.; Zhou, S.; and Liu, X. 2023.
\newblock Learn from relational correlations and periodic events for temporal knowledge graph reasoning.
\newblock In \emph{Proceedings of the 46th international ACM SIGIR conference on research and development in information retrieval}, 1559--1568.

\bibitem[{Liu et~al.(2024{\natexlab{a}})Liu, Yin, Liu, and Tian}]{liu2024reinforcement}
Liu, R.; Yin, G.; Liu, Z.; and Tian, Y. 2024{\natexlab{a}}.
\newblock Reinforcement learning with time intervals for temporal knowledge graph reasoning.
\newblock \emph{Information Systems}, 120: 102292.

\bibitem[{Liu et~al.(2024{\natexlab{b}})Liu, Zhang, Ma, Liang, Xu, and Zong}]{liu2024temporal}
Liu, X.; Zhang, J.; Ma, C.; Liang, W.; Xu, B.; and Zong, L. 2024{\natexlab{b}}.
\newblock Temporal Knowledge Graph Reasoning with Dynamic Hypergraph Embedding.
\newblock In \emph{Proceedings of the 2024 Joint International Conference on Computational Linguistics, Language Resources and Evaluation (LREC-COLING 2024)}, 15742--15751.

\bibitem[{Lv et~al.(2024)Lv, Huang, Ouyang, Chen, and Xie}]{lv2024rlgnet}
Lv, A.; Huang, Y.; Ouyang, G.; Chen, Y.; and Xie, H. 2024.
\newblock RLGNet: Repeating-Local-Global History Network for Temporal Knowledge Graph Reasoning.
\newblock \emph{arXiv preprint arXiv:2404.00586}.

\bibitem[{Mahdisoltani, Biega, and Suchanek(2013)}]{mahdisoltani2013yago3}
Mahdisoltani, F.; Biega, J.; and Suchanek, F.~M. 2013.
\newblock Yago3: A knowledge base from multilingual wikipedias.
\newblock In \emph{CIDR}.

\bibitem[{Ou and Jian(2024)}]{ou2024effective}
Ou, Y.; and Jian, P. 2024.
\newblock Effective Integration of Text Diffusion and Pre-Trained Language Models with Linguistic Easy-First Schedule.
\newblock In \emph{Proceedings of the 2024 Joint International Conference on Computational Linguistics, Language Resources and Evaluation (LREC-COLING 2024)}, 5551--5561.

\bibitem[{Pareja et~al.(2020)Pareja, Domeniconi, Chen, Ma, Suzumura, Kanezashi, Kaler, Schardl, and Leiserson}]{pareja2020evolvegcn}
Pareja, A.; Domeniconi, G.; Chen, J.; Ma, T.; Suzumura, T.; Kanezashi, H.; Kaler, T.; Schardl, T.; and Leiserson, C. 2020.
\newblock Evolvegcn: Evolving graph convolutional networks for dynamic graphs.
\newblock In \emph{Proceedings of the AAAI conference on artificial intelligence}, volume~34, 5363--5370.

\bibitem[{Park et~al.(2022)Park, Liu, Mehta, Cristofor, Faloutsos, and Dong}]{park2022evokg}
Park, N.; Liu, F.; Mehta, P.; Cristofor, D.; Faloutsos, C.; and Dong, Y. 2022.
\newblock Evokg: Jointly modeling event time and network structure for reasoning over temporal knowledge graphs.
\newblock In \emph{Proceedings of the fifteenth ACM international conference on web search and data mining}, 794--803.

\bibitem[{Ramesh et~al.(2022)Ramesh, Dhariwal, Nichol, Chu, and Chen}]{ramesh2022hierarchical}
Ramesh, A.; Dhariwal, P.; Nichol, A.; Chu, C.; and Chen, M. 2022.
\newblock Hierarchical text-conditional image generation with clip latents.
\newblock \emph{arXiv preprint arXiv:2204.06125}, 1(2): 3.

\bibitem[{Rombach et~al.(2022)Rombach, Blattmann, Lorenz, Esser, and Ommer}]{rombach2022high}
Rombach, R.; Blattmann, A.; Lorenz, D.; Esser, P.; and Ommer, B. 2022.
\newblock High-resolution image synthesis with latent diffusion models.
\newblock In \emph{Proceedings of the IEEE/CVF conference on computer vision and pattern recognition}, 10684--10695.

\bibitem[{Saharia et~al.(2022)Saharia, Chan, Saxena, Li, Whang, Denton, Ghasemipour, Gontijo~Lopes, Karagol~Ayan, Salimans et~al.}]{saharia2022photorealistic}
Saharia, C.; Chan, W.; Saxena, S.; Li, L.; Whang, J.; Denton, E.~L.; Ghasemipour, K.; Gontijo~Lopes, R.; Karagol~Ayan, B.; Salimans, T.; et~al. 2022.
\newblock Photorealistic text-to-image diffusion models with deep language understanding.
\newblock \emph{Advances in neural information processing systems}, 35: 36479--36494.

\bibitem[{Schlichtkrull et~al.(2018)Schlichtkrull, Kipf, Bloem, Van Den~Berg, Titov, and Welling}]{schlichtkrull2018modeling}
Schlichtkrull, M.; Kipf, T.~N.; Bloem, P.; Van Den~Berg, R.; Titov, I.; and Welling, M. 2018.
\newblock Modeling relational data with graph convolutional networks.
\newblock In \emph{The semantic web: 15th international conference, ESWC 2018, Heraklion, Crete, Greece, June 3--7, 2018, proceedings 15}, 593--607. Springer.

\bibitem[{Song, Meng, and Ermon(2020)}]{song2020denoising}
Song, J.; Meng, C.; and Ermon, S. 2020.
\newblock Denoising diffusion implicit models.
\newblock \emph{arXiv preprint arXiv:2010.02502}.

\bibitem[{Strudel et~al.(2022)Strudel, Tallec, Altch{\'e}, Du, Ganin, Mensch, Grathwohl, Savinov, Dieleman, Sifre et~al.}]{strudel2022self}
Strudel, R.; Tallec, C.; Altch{\'e}, F.; Du, Y.; Ganin, Y.; Mensch, A.; Grathwohl, W.; Savinov, N.; Dieleman, S.; Sifre, L.; et~al. 2022.
\newblock Self-conditioned embedding diffusion for text generation.
\newblock \emph{arXiv preprint arXiv:2211.04236}.

\bibitem[{Trivedi et~al.(2017)Trivedi, Dai, Wang, and Song}]{trivedi2017know}
Trivedi, R.; Dai, H.; Wang, Y.; and Song, L. 2017.
\newblock Know-evolve: Deep temporal reasoning for dynamic knowledge graphs.
\newblock In \emph{international conference on machine learning}, 3462--3471. PMLR.

\bibitem[{Trouillon et~al.(2016)Trouillon, Welbl, Riedel, Gaussier, and Bouchard}]{trouillon2016complex}
Trouillon, T.; Welbl, J.; Riedel, S.; Gaussier, {\'E}.; and Bouchard, G. 2016.
\newblock Complex embeddings for simple link prediction.
\newblock In \emph{International conference on machine learning}, 2071--2080. PMLR.

\bibitem[{Wang, Han, and Poon(2023)}]{wang2023re}
Wang, K.; Han, S.~C.; and Poon, J. 2023.
\newblock Re-Temp: Relation-Aware Temporal Representation Learning for Temporal Knowledge Graph Completion.
\newblock \emph{arXiv preprint arXiv:2310.15722}.

\bibitem[{Wang et~al.(2020)Wang, Li, Wang, Parulian, Han, Ma, Tu, Lin, Zhang, Liu et~al.}]{Wang-2020covid}
Wang, Q.; Li, M.; Wang, X.; Parulian, N.; Han, G.; Ma, J.; Tu, J.; Lin, Y.; Zhang, H.; Liu, W.; et~al. 2020.
\newblock COVID-19 literature knowledge graph construction and drug repurposing report generation.
\newblock \emph{arXiv preprint arXiv:2007.00576}.

\bibitem[{Wang et~al.(2017)Wang, Mao, Wang, and Guo}]{wang2017knowledge}
Wang, Q.; Mao, Z.; Wang, B.; and Guo, L. 2017.
\newblock Knowledge graph embedding: A survey of approaches and applications.
\newblock \emph{IEEE transactions on knowledge and data engineering}, 29(12): 2724--2743.

\bibitem[{Wu et~al.(2020)Wu, Cao, Cheung, and Hamilton}]{wu2020temp}
Wu, J.; Cao, M.; Cheung, J. C.~K.; and Hamilton, W.~L. 2020.
\newblock Temp: Temporal message passing for temporal knowledge graph completion.
\newblock \emph{arXiv preprint arXiv:2010.03526}.

\bibitem[{Xie et~al.(2023)Xie, Zhu, Liu, Zhou, and Huang}]{xie2023targat}
Xie, Z.; Zhu, R.; Liu, J.; Zhou, G.; and Huang, J.~X. 2023.
\newblock TARGAT: A time-aware relational graph attention model for temporal knowledge graph embedding.
\newblock \emph{IEEE/ACM Transactions on Audio, Speech, and Language Processing}, 31: 2246--2258.

\bibitem[{Xu et~al.(2023)Xu, Ou, Xu, and Fu}]{xu2023temporal}
Xu, Y.; Ou, J.; Xu, H.; and Fu, L. 2023.
\newblock Temporal knowledge graph reasoning with historical contrastive learning.
\newblock In \emph{Proceedings of the AAAI Conference on Artificial Intelligence}, volume~37, 4765--4773.

\bibitem[{Xuan et~al.(2023)Xuan, Liu, Li, and Yin}]{xuan2023knowledge}
Xuan, H.; Liu, Y.; Li, B.; and Yin, H. 2023.
\newblock Knowledge enhancement for contrastive multi-behavior recommendation.
\newblock In \emph{Proceedings of the sixteenth ACM international conference on web search and data mining}, 195--203.

\bibitem[{Yin and Cui(2016)}]{yin2016spatio}
Yin, H.; and Cui, B. 2016.
\newblock \emph{Spatio-temporal recommendation in social media}.
\newblock Springer.

\bibitem[{Yu et~al.(2022)Yu, Yin, Xia, Chen, Cui, and Nguyen}]{yu2022graph}
Yu, J.; Yin, H.; Xia, X.; Chen, T.; Cui, L.; and Nguyen, Q. V.~H. 2022.
\newblock Are graph augmentations necessary? simple graph contrastive learning for recommendation.
\newblock In \emph{Proceedings of the 45th international ACM SIGIR conference on research and development in information retrieval}, 1294--1303.

\bibitem[{Z et~al.(2020)Z, P., Y., and V.}]{han2020targat}
Z, H.; P., C.; Y., M.; and V., T. 2020.
\newblock Explainable subgraph reasoning for forecasting on temporal knowledge graphs.
\newblock \emph{In International Conference on Learning Represen- tations}.

\bibitem[{Zhang et~al.(2024)Zhang, Hui, Mu, Sun, and Tian}]{zhang2024historically}
Zhang, J.; Hui, B.; Mu, C.; Sun, M.; and Tian, L. 2024.
\newblock Historically Relevant Event Structuring for Temporal Knowledge Graph Reasoning.
\newblock \emph{arXiv preprint arXiv:2405.10621}.

\bibitem[{Zhang and Zhou(2022)}]{zhang2022temporal}
Zhang, L.; and Zhou, D. 2022.
\newblock Temporal knowledge graph completion with approximated Gaussian process embedding.
\newblock In \emph{Proceedings of the 29th International Conference on Computational Linguistics}, 4697--4706.

\bibitem[{Zhou et~al.(2023)Zhou, Kang, Tian, and Su}]{zhou2023intensity}
Zhou, W.-T.; Kang, Z.; Tian, L.; and Su, Y. 2023.
\newblock Intensity-free convolutional temporal point process: Incorporating local and global event contexts.
\newblock \emph{Information Sciences}, 646: 119318.

\bibitem[{Zhou et~al.(2022)Zhou, Chen, He, Ye, and Sun}]{zhou2022re}
Zhou, Y.; Chen, X.; He, B.; Ye, Z.; and Sun, L. 2022.
\newblock Re-thinking knowledge graph completion evaluation from an information retrieval perspective.
\newblock In \emph{Proceedings of the 45th International ACM SIGIR Conference on Research and Development in Information Retrieval}, 916--926.

\bibitem[{Zhu et~al.(2021)Zhu, Chen, Fan, Cheng, and Zhang}]{zhu2021learning}
Zhu, C.; Chen, M.; Fan, C.; Cheng, G.; and Zhang, Y. 2021.
\newblock Learning from history: Modeling temporal knowledge graphs with sequential copy-generation networks.
\newblock In \emph{Proceedings of the AAAI conference on artificial intelligence}, volume~35, 4732--4740.

\end{thebibliography}

\end{document}